\pdfoutput=1

\documentclass[11pt]{article}

\usepackage[]{ACL2023}

\usepackage{times}
\usepackage{latexsym}

\usepackage[T1]{fontenc}

\usepackage[utf8]{inputenc}

\usepackage{microtype}

\usepackage{inconsolata}

\usepackage{xurl}

\usepackage{booktabs}
\usepackage{xcolor, colortbl}
\usepackage{xfp} 
\usepackage{pgf}
\usepackage{tikz}
\usepackage{amsmath}
\usepackage{amssymb}
\usepackage{enumitem}
\usepackage{multicol} 

\usepackage[most]{tcolorbox}
\usepackage{listings}
\usepackage{fvextra}
\usepackage{txfonts}

\lstdefinestyle{promptstyle}{
    basicstyle=\ttfamily,
    numbers=left,
    numberstyle=\color{gray},
    numbersep=12pt,
    breaklines=true,
    columns=fullflexible,
    frame=none,
    showstringspaces=false
}

\newif\ifshowcomments
\showcommentstrue 

\title{Typological Feature Prediction with Large Language Models: \\An In-Context Learning Approach}

\author{
 \textbf{Qianwen Wang$^{\varheartsuit\hspace{1pt}*}$} \,
 \textbf{York Hay Ng$^{\varheartsuit\hspace{1pt}*}$} \,
 \textbf{Aditya Khan$^{\varheartsuit}$} \,
 \textbf{En-Shiun Annie Lee$^{\varheartsuit\hspace{1pt}\clubsuit}$}
\\
 $^{\varheartsuit}$University of Toronto, Canada\quad
 $^{\clubsuit}$Ontario Tech University, Canada\\
 \\
 \texttt{vivianqianwen.wang@mail.utoronto.ca,}\\
 \texttt{york.ng@mila.quebec, aditya.khan@columbia.edu}
}

\begin{document}
\maketitle
\begingroup
\renewcommand\thefootnote{\fnsymbol{footnote}}
\footnotetext[1]{The authors contributed equally.}
\endgroup
\begin{abstract}
Typological features are widely used in multilingual NLP, and the prediction of such features holds downstream utility. However, existing methods to predict missing values lack interpretable justifications for predictions, while their performance across resource levels and feature types remains underexplored. Given LLMs' abilities in meta-linguistic reasoning and in providing rationales, we investigate LLMs' performance in typological feature prediction via an in-context learning approach with linguistic data from URIEL+ and Glottolog. We find that zero-shot prompting is insufficient, but when given phylogenetic and geographic neighbour evidence, LLMs substantially outperform all baselines without disadvantaging low-resource languages. We further find that most LLM rationales are consistent with the provided evidence, offering a step toward explainable typological feature prediction. 
\end{abstract}

\section{Introduction}
\label{sec:intro}

Typological features describe the structural properties of languages. They are widely used across multilingual NLP\@ for tasks including transfer language selection
\citep{lin2019choosing}, cross-lingual dependency parsing
\citep{naseem-etal-2012-selective,tackstrom2013target,
de2018parameter,ustun2020udapter},
and performance prediction of multilingual models \citep{hirak-etal-2026-assessing, anugraha2025proxylm,xia-etal-2020-predicting}. Surveys have confirmed that typological features benefit downstream performance 
\citep{ponti2019modeling,bjerva-2024-role}.
 
The databases that house these features, however, are usually incomplete. URIEL+ \citep{khan2025uriel+}, the most recent typological knowledge base, aggregates features from databases such as WALS \citep{wals}, Grambank \citep{grambank} and PHOIBLE \citep{phoible}, yet 87\% of its typological feature matrix remains empty \citep{ng2025less}. Moreover, sparsity is worst precisely for the low-resource languages (LRLs) that stand to benefit most from methods using those typological features \citep{shipton-etal-2026-simple}.

Typological feature prediction is a task aimed at addressing this gap. Existing approaches use $k$-nearest-neighbour imputation \citep{littell2017uriel}, low-rank matrix completion \citep{khan2025uriel+, ng-etal-2026-modality}, and random forests trained on external features \citep{amirzadeh2025data2lang2vec}. However, inequalities in prediction performance across resource levels and feature types remain underexplored, while output predictions are usually not accompanied with interpretable justifications.

Recent work has shown that LLMs exhibit meta-linguistic reasoning skills \citep{yang2025linggym}. Coupled with their ability to output rationales, they are a natural candidate for this task. In this work, we ask: \textit{how well do pretrained LLMs reason over linguistic evidence, and provide rationales, for typological feature prediction?}

We propose an in-context learning approach, constructing prompts with data from URIEL+ \cite{khan2025uriel+} and Glottolog \cite{Glottolog}, a catalogue of the world's languages and families. Our exploration suggests that:
\begin{enumerate}[]
    \item Zero-shot prompting, where only language metadata is provided, is insufficient for this task, but LLMs excel at reasoning over linguistic evidence when evidence is provided.
    \item LLMs, once given sufficient contextual information, exhibit consistent performance across language resource levels and typological properties, consistently exceeding non-LLM baselines on typological feature prediction.
    \item Human annotators confirm that our prompt design supplies sufficient evidence, and find most LLM generated rationales to be consistent with the evidence.
\end{enumerate}

\section{Related Work}
\label{sec:related}
 
\paragraph{Typological Feature Prediction.}
The SIGTYP 2020 Shared Task \citep{bjerva2020sigtyp} formalized the task of typological feature prediction over WALS. The winning system combined conditional probabilities with language embeddings \citep{vastl2020predicting}. Methods that operate on typological data alone include $k$-nearest-neighbour imputation and SoftImpute, a low-rank matrix factorization used in URIEL+ \citep{khan2025uriel+} that remains a strong baseline.

A separate line of work incorporates external data.  \citet{amirzadeh2025data2lang2vec} proposed per-feature random forests trained with POS-tag distributions and Wikipedia statistics, while prior probing-based approaches predict features from multilingual representations trained on parallel text \citep{malaviya2017learning,ostling2023language}.

\citet{bjerva-2024-role} critiques these methods as fundamentally correlation-based. Easy features are trivially predicted while performance in atypical cases, despite being the most typologically interesting, remains poor. In a survey of linguists, explainability was rated as the most important property of prediction methods, which all existing methods lack. Moreover, methods that rely on external data sources or parallel text are inapplicable to the low-resource languages with more missing features.

 
\paragraph{LLMs and Linguistic Reasoning.}
A growing body of work demonstrates that LLMs can perform substantive linguistic analysis. \citet{tanzer2024benchmark} showed that prompting an LLM with a single grammar book enables translation of unseen low-resource languages, although \citet{aycock2025can} found that performance mainly hinges on the book's parallel examples. Meanwhile, \citet{yang2025linggym} found that linguistic cues (e.g. grammatical descriptions) consistently improve LLMs' meta-linguistic reasoning across typologically diverse languages. In typological feature prediction, \citet{hus2026rag} applied retrieval-augmented generation with grammar books to predict Grambank features, demonstrating that retrieval yielded substantial improvement.

Our work continues to investigate LLMs' ability to reason over linguistic evidence. In contrast to prior approaches, our method does not rely on external resources where availability is limited for low-resource languages, while aiding interpretability by providing rationales of predictions.

\begin{table*}[t]
\centering
\small
\setlength{\tabcolsep}{6pt}
\begin{tabular}{l c c c c @{\hspace{1.5em}}c c c c @{\hspace{1.5em}}c}
\toprule
\textbf{Method} & \textbf{Inputs} & \textbf{LRL} & \textbf{MRL} & \textbf{HRL} & \textbf{S} & \textbf{P} & \textbf{INV} & \textbf{M} & \textbf{Overall} \\
\midrule

\quad Random
& --
& \cellcolor{blue!2}.500
& \cellcolor{blue!11}.613
& \cellcolor{blue!2}.584
& \cellcolor{blue!0}.446
& \cellcolor{blue!12}.689
& \cellcolor{blue!21}.696
& \cellcolor{blue!0}.402
& \cellcolor{blue!5}.568 \\

\quad Majority
& --
& \cellcolor{blue!12}.637
& \cellcolor{blue!15}.660
& \cellcolor{blue!8}.659
& \cellcolor{blue!6}.565
& \cellcolor{blue!22}.777
& \cellcolor{blue!23}.730
& \cellcolor{blue!1}.427
& \cellcolor{blue!12}.653 \\

\quad kNN (cosine) ($k=3$)
& (A)
& \cellcolor{blue!24}.754
& \cellcolor{blue!24}.741
& \cellcolor{blue!27}.817
& \cellcolor{blue!16}.690
& \cellcolor{blue!28}.815
& \cellcolor{blue!43}.927
& \cellcolor{blue!19}.662
& \cellcolor{blue!26}.775 \\

\quad SoftImpute
& (A)
& \cellcolor{blue!30}.800
& \cellcolor{blue!33}.817
& \cellcolor{blue!31}.841
& \cellcolor{blue!20}.721
& \cellcolor{blue!42}.907
& \cellcolor{blue!41}.908
& \cellcolor{blue!25}.715
& \cellcolor{blue!32}.821 \\

\quad kNN (phylogenetic) ($k=6$)
& (A)\,+\,(B)
& \cellcolor{blue!22}.732
& \cellcolor{blue!22}.718
& \cellcolor{blue!12}.695
& \cellcolor{blue!13}.659
& \cellcolor{blue!25}.794
& \cellcolor{blue!26}.753
& \cellcolor{blue!12}.600
& \cellcolor{blue!19}.714 \\

\quad kNN (geographic) ($k=6$)
& (A)\,+\,(C)
& \cellcolor{blue!20}.720
& \cellcolor{blue!28}.773
& \cellcolor{blue!14}.710
& \cellcolor{blue!16}.689
& \cellcolor{blue!25}.795
& \cellcolor{blue!28}.774
& \cellcolor{blue!17}.646
& \cellcolor{blue!21}.733 \\

\quad Random Forest
& (A)\,+\,(B)\,+\,(C)
& \cellcolor{blue!22}.739
& \cellcolor{blue!22}.719
& \cellcolor{blue!16}.734
& \cellcolor{blue!14}.663
& \cellcolor{blue!33}.852
& \cellcolor{blue!25}.743
& \cellcolor{blue!12}.595
& \cellcolor{blue!21}.731 \\

\addlinespace
\multicolumn{10}{l}{\textit{Llama-3.1-70B in-context prompting}} \\

\quad Base, zero-shot (metadata only)
& --
& \cellcolor{blue!0}.433
& \cellcolor{blue!0}.424
& \cellcolor{blue!0}.531
& \cellcolor{blue!0}.463
& \cellcolor{blue!0}.521
& \cellcolor{blue!2}.418
& \cellcolor{blue!1}.442
& \cellcolor{blue!0}.465 \\

\quad \,+\,anchor features
& (A)
& \cellcolor{blue!1}.459
& \cellcolor{blue!2}.482
& \cellcolor{blue!2}.587
& \cellcolor{blue!1}.483
& \cellcolor{blue!3}.592
& \cellcolor{blue!9}.547
& \cellcolor{blue!0}.411
& \cellcolor{blue!2}.512 \\

\quad \,+\,anchor\,+\,phylogenetic neighb.
& (A)\,+\,(B)
& \cellcolor{blue!25}.759
& \cellcolor{blue!30}.795
& \cellcolor{blue!23}.789
& \cellcolor{blue!27}.785
& \cellcolor{blue!32}.845
& \cellcolor{blue!29}.787
& \cellcolor{blue!22}.690
& \cellcolor{blue!27}.782 \\

\quad \,+\,anchor\,+\,geographic neighb.
& (A)\,+\,(C)
& \cellcolor{blue!22}.736
& \cellcolor{blue!28}.778
& \cellcolor{blue!30}.833
& \cellcolor{blue!28}.793
& \cellcolor{blue!23}.780
& \cellcolor{blue!37}.867
& \cellcolor{blue!25}.720
& \cellcolor{blue!28}.788 \\

\quad \,+\,all
& (A)\,+\,(B)\,+\,(C)
& \cellcolor{blue!45}\textbf{.919}
& \cellcolor{blue!44}.895
& \cellcolor{blue!42}.914
& \cellcolor{blue!44}.914
& \cellcolor{blue!43}.917
& \cellcolor{blue!44}.933
& \cellcolor{blue!45}\textbf{.868}
& \cellcolor{blue!45}\textbf{.909} \\

\addlinespace
\multicolumn{10}{l}{\textit{Gemma-4-31B in-context prompting}} \\

\quad Base, zero-shot (metadata only)
& --
& \cellcolor{blue!1}.472
& \cellcolor{blue!6}.547
& \cellcolor{blue!0}.539
& \cellcolor{blue!2}.507
& \cellcolor{blue!10}.669
& \cellcolor{blue!0}.336
& \cellcolor{blue!1}.450
& \cellcolor{blue!2}.521 \\

\quad \,+\,anchor features
& (A)
& \cellcolor{blue!7}.580
& \cellcolor{blue!15}.654
& \cellcolor{blue!5}.629
& \cellcolor{blue!2}.508
& \cellcolor{blue!22}.777
& \cellcolor{blue!19}.677
& \cellcolor{blue!3}.476
& \cellcolor{blue!9}.622 \\

\quad \,+\,anchor\,+\,phylogenetic neighb.
& (A)\,+\,(B)
& \cellcolor{blue!33}.828
& \cellcolor{blue!37}.847
& \cellcolor{blue!29}.831
& \cellcolor{blue!27}.786
& \cellcolor{blue!40}.899
& \cellcolor{blue!37}.872
& \cellcolor{blue!31}.762
& \cellcolor{blue!34}.835 \\

\quad \,+\,anchor\,+\,geographic neighb.
& (A)\,+\,(C)
& \cellcolor{blue!37}.858
& \cellcolor{blue!41}.878
& \cellcolor{blue!45}\textbf{.929}
& \cellcolor{blue!38}.871
& \cellcolor{blue!43}.914
& \cellcolor{blue!45}\textbf{.944}
& \cellcolor{blue!41}.841
& \cellcolor{blue!43}.893 \\

\quad \,+\,all
& (A)\,+\,(B)\,+\,(C)
& \cellcolor{blue!34}.834
& \cellcolor{blue!36}.840
& \cellcolor{blue!41}.907
& \cellcolor{blue!34}.841
& \cellcolor{blue!40}.900
& \cellcolor{blue!41}.909
& \cellcolor{blue!36}.802
& \cellcolor{blue!38}.865 \\

\addlinespace
\multicolumn{10}{l}{\textit{GPT-5.5 in-context prompting}} \\

\quad Base, zero-shot (metadata only)
& --
& \cellcolor{blue!11}.622
& \cellcolor{blue!15}.657
& \cellcolor{blue!12}.693
& \cellcolor{blue!9}.607
& \cellcolor{blue!21}.767
& \cellcolor{blue!17}.653
& \cellcolor{blue!13}.602
& \cellcolor{blue!13}.661 \\

\quad \,+\,anchor features
& (A)
& \cellcolor{blue!17}.687
& \cellcolor{blue!20}.707
& \cellcolor{blue!19}.754
& \cellcolor{blue!8}.600
& \cellcolor{blue!33}.851
& \cellcolor{blue!32}.816
& \cellcolor{blue!11}.586
& \cellcolor{blue!19}.719 \\

\quad \,+\,anchor\,+\,phylogenetic neighb.
& (A)\,+\,(B)
& \cellcolor{blue!35}.847
& \cellcolor{blue!40}.872
& \cellcolor{blue!30}.837
& \cellcolor{blue!30}.811
& \cellcolor{blue!42}.908
& \cellcolor{blue!42}.921
& \cellcolor{blue!28}.742
& \cellcolor{blue!36}.851 \\

\quad \,+\,anchor\,+\,geographic neighb.
& (A)\,+\,(C)
& \cellcolor{blue!40}.880
& \cellcolor{blue!44}.899
& \cellcolor{blue!43}.915
& \cellcolor{blue!45}\textbf{.923}
& \cellcolor{blue!45}\textbf{.928}
& \cellcolor{blue!44}.937
& \cellcolor{blue!36}.800
& \cellcolor{blue!44}.900 \\

\quad \,+\,all
& (A)\,+\,(B)\,+\,(C)
& \cellcolor{blue!38}.869
& \cellcolor{blue!45}\textbf{.905}
& \cellcolor{blue!37}.881
& \cellcolor{blue!38}.872
& \cellcolor{blue!43}.917
& \cellcolor{blue!43}.930
& \cellcolor{blue!36}.807
& \cellcolor{blue!41}.885 \\

\bottomrule
\end{tabular}
\caption{Macro F1 of feature prediction methods on URIEL+, grouped by language resource level (LRL/MRL/HRL) and by feature type (S=syntax, P=phonology, INV=phonetic inventory, M=morphology). \textbf{Inputs}: (A)=typological matrix, (B)=phylogenetic neighbours, (C)=geographic neighbours. Cell shading is column-wise (darker = higher F1); per-column best is in \textbf{bold}.}
\label{tab:main-results}
\end{table*}

\section{Method}
\label{sec:method}

\subsection{Task Formulation}

We frame typological feature prediction as an imputation task over an existing knowledge base against held-out observed values, following the formulation identified in \citet{bjerva-2024-role}. We apply this task on the URIEL+ typological matrix \citep{khan2025uriel+}, given its comprehensive coverage of typological features and languages. Let $\mathbf{M} \in \{0, 1, \bot\}^{N \times F}$ denote URIEL+, with $N{=}4{,}555$ languages and $F{=}800$ binary typological features, where $\bot$ indicates a missing entry. Given a query $(\ell, f)$ over target language $\ell$ and typological feature $f$, where $\mathbf{M}[\ell, f] = \bot$, the task is to predict $\hat{v} \in \{0, 1\}$. We use Glottolog \citep{Glottolog} as the sole source of metadata for $\ell$, as it provides data for all languages covered by URIEL+.

\paragraph{Data splits.}
We aim to preserve equal representation between typological feature types and language resource levels in our data splits. URIEL+ documents four types of typological features: syntactic, morphological, phonological and inventory. Moreover, we categorise the resource level of languages by the number of features with an observed value. The bottom 33\% of languages form the low-resource group (LRL), the middle 34\% the medium-resource group (MRL), and the top 33\% the high-resource group (HRL).

We then sample 100 $(\ell, f)$ observed pairs for each resource level and feature type where $\mathbf{M}[\ell, f]\neq\bot$, yielding $1{,}200$ pairs as our test set \footnote{Further statistics on the composition of the test set are provided in Appendix~\ref{app:test-composition}.}. A separate validation set of equal size is held out for hyperparameter tuning for baseline methods. The remaining 888,741 observed entries form the training matrix $\mathbf{M}_{\text{train}}$.


\subsection{Prompt Construction}
\label{sec:prompt}

For each query, we construct a prompt from four content blocks drawn from URIEL+ and Glottolog. The full prompt template is shown in Appendix~\ref{app:prompt-template}.

\paragraph{(1) Target language metadata.} The language's name, Glottocode, ISO 639-3 code, language family, and macroarea.

\paragraph{(2) Anchor features.} For each target feature $f$, we pre-compute the top 10 features most correlated with $f$ across $\mathbf{M}_{\text{train}}$, termed \textit{anchor features}. Given a query language $\ell$, we then provide the top 5 anchor feature values that are observed for $\ell$
. Additionally, for each observed anchor feature $a$ with value $v_a$, we provide the global prevalence of the feature $P(f{=}1 \mid a{=}v_a)$ over $\mathbf{M}_{\text{train}}$.

\paragraph{(3) Phylogenetic neighbours.} We build a pool of candidates by breadth-first search from $\ell$ in the Glottolog family tree, selecting 5 languages greedily to maximise coverage of the target and anchor features. For each neighbour we include its value for $f$ (if observed), up to 3 observed anchor values, and its proximity rank. We further aggregate \emph{yes/no/missing} vote counts for $f$ across the set of neighbours, and provide details of the closest phylogenetic neighbour supporting both values of $f$ as contrastive evidence.

\paragraph{(4) Geographic neighbours.} We construct this pool identically as (3), with the candidate pool ranked by Haversine distance over Glottolog coordinates. We take the same per-neighbour data, vote counts, and contrastive evidence as above.

\subsection{Baselines}
\label{sec:baselines}

We compare against seven imputation methods, organized by the information they consume:

\paragraph{Trivial baselines.}
\texttt{Random} draws each prediction from a Bernoulli distribution with the per-feature positive rate over $\mathbf{M}_\text{train}$. \textbf{Majority} predicts the more frequent value per feature.

\paragraph{Matrix-only methods.}
\texttt{kNN-cosine} \citep{littell2017uriel} performs $k$-nearest-neighbour imputation over $\mathbf{M}_\text{train}$ using cosine similarity. \texttt{SoftImpute} \citep{mazumder2010spectral} completes the matrix via iterative low-rank completion with soft-thresholded SVD, and it was the strongest baseline in URIEL+  \citep{khan2025uriel+}.

\paragraph{Glottolog-based kNN.}
To test the utility of phylogenetic and geographic information, we evaluate two variants of \texttt{kNN-cosine}. \texttt{kNN-phylogenetic} ranks neighbours by path distance in the Glottolog tree. \texttt{kNN-geographic} ranks them by Haversine distance over geographic coordinates.

\paragraph{Random Forest.}
For a direct non-LLM comparison, we mirror \citet{amirzadeh2025data2lang2vec}'s architecture but restrict the model to the data available to the LLM. For each feature, we train a random forest classifier\footnote{Input features and hyperparameter choices are detailed in Appendix~\ref{app:hparams}.} over target language metadata (family, macroarea, coordinates), the values and global prevalence of anchor features, as well as vote count for phylogenetic and geographic neighbours. Since this baseline is provided the same information from Glottolog as the LLM, we aim to isolate the impact of LLM reasoning abilities.
 
\section{Results}
\label{sec:results}
We evaluate on two open-source models, Llama-3.1-70B \citep{grattafiori2024llama} and Gemma-4-31B \cite{gemma4}, on their instruction-tuned variants, alongside one proprietary model, GPT-5.5 \citep{openai2026gpt55}. Table~\ref{tab:main-results} reports macro F1 on the test set across resource levels and feature types.

\paragraph{Zero-shot prompting underperforms baselines.}
All three LLMs underperform the strongest baselines when given only language metadata (Llama: F1 0.465, Gemma: F1 0.521, GPT-5.5: F1 0.661), where they effectively perform zero-shot prediction. Adding anchor features provides a moderate improvement for all three models, reaching F1 0.512, 0.622, and 0.719 respectively. Nonetheless, these configurations remain below \texttt{kNN-cosine} (F1 0.775) and \texttt{SoftImpute} (F1 0.821). This highlights that metadata-only prompting, where LLMs have little evidence to reason over, is insufficient, therefore motivating the inclusion of typological and neighbour evidence.

\paragraph{LLMs excel at reasoning over evidence.}
When neighbour evidence is introduced, performance improves substantially. The strongest neighbour-based configuration of every LLM outperforms \texttt{SoftImpute}, the best performing baseline (F1 0.821). Llama with all inputs achieves the highest overall performance (F1 0.909). On the other hand, Gemma and GPT-5.5 instead perform best when anchor features and geographic neighbour evidence are included, reaching F1 0.893 and 0.900 respectively. This shows that providing linguistic evidence improves LLMs' prediction performance.

\texttt{Random Forest} receives the same Glottolog-derived inputs as the LLMs, yet it scores below even baselines such as \texttt{SoftImpute} and \texttt{kNN-cosine}, which see only the typological matrix. This suggests that reasoning ability is a key source of the LLMs' advantage. Appendix \ref{app:plurality} further demonstrates that LLMs' reasoning exceeds simply following the plurality, and that performance is robust even when evidence disagrees.

\paragraph{With sufficient evidence, performance gaps narrow across resource levels and feature types.}
With metadata alone, all three LLMs perform noticeably worse on LRLs than HRLs. Conversely, when given all inputs, Llama exhibits marginally better performance on LRLs (F1 0.919) than HRLs (F1 0.914). While the inequality persists in Gemma, the performance gap between HRLs and LRLs remains similar, from 0.067 (base) to 0.073 (all). For GPT-5.5, this gap narrows substantially from 0.071 (base) to 0.012 (all).

Across feature types, most baseline methods exhibit varying performance, generally favouring phonological and phonetic inventory features above syntactic and morphological features. By comparison, the LLMs are more consistent across types: when given all inputs, Llama and Gemma's mean performance across feature types ranges only 0.065 and 0.107 respectively. For GPT-5.5, the corresponding range is 0.123.

 


\section{Annotation Study}
\label{sec:human}
 
To study whether the prompt provides sufficient evidence, and whether LLM generated rationales are consistent with that evidence, we conducted a human annotation study over a subset of 180 $(\ell, f)$ pairs, collecting 15 pairs per resource level and feature type. Five annotators, given the same prompt as the LLM, independently recorded their prediction and confidence. Separately, they
annotated whether the LLM's rationale contradicts the evidence provided. The full protocol is described in Appendix 
\ref{app:annotation-protocol}.
 
\paragraph{Sufficiency of evidence in prompts.}
We observed high inter-annotator agreement on predictions, with Fleiss' $\kappa = 0.888$. Aggregated by majority vote, human predictions achieved F1 0.987, with individual annotators averaging an F1 of 0.963. This confirms that the prompt provides sufficient evidence to perform feature prediction.
 
\paragraph{LLM-human comparison.}
On the same subset, our best performing LLM, Llama-3.1-70B, achieves F1 0.915. While performance is below that of humans, there is strong agreement between Llama and aggregated human predictions, exhibiting Cohen's $\kappa = 0.831$.
 
\paragraph{Confidence calibration.}
Table~\ref{tab:confidence} reports the accuracy for humans and Llama, grouped by confidence. We observe that for both humans and Llama, the accuracy of predictions consistently increases with confidence, with Llama achieving perfect accuracy when high confidence is reported. While Llama tends to report lower confidence compared to humans overall, this result highlights the validity of Llama's reported confidences.
 
\begin{table}[t]
\centering
\small
\begin{tabular}{l c c c c}
\toprule
& \multicolumn{2}{c}{\textbf{Human}} & \multicolumn{2}{c}{\textbf{LLM}} \\
\cmidrule(lr){2-3} \cmidrule(lr){4-5}
\textbf{Confidence} & \textbf{Acc.} & \textbf{\%} & \textbf{Acc.} & \textbf{\%} \\
\midrule
Low    & 0.794 & 0.134 & 0.643 & 0.156 \\
Medium & 0.963 & 0.256 & 0.962 & 0.439 \\
High   & 0.997 & 0.610 & 1.000 & 0.406 \\
\bottomrule
\end{tabular}
\caption{Accuracy by self-reported confidence tier.
\textbf{\%} indicates the proportion of predictions in each tier.
}
\label{tab:confidence}
\end{table}
 
\paragraph{Rationale quality.}
On average, annotators identified contradictions in 20.9\% of Llama's rationales,
though inter-annotator agreement on contradiction judgments was
moderate (Fleiss' $\kappa = 0.415$), which reflects the subjective
nature of evaluating rationales. While the fact that the majority of rationales are consistent with the provided evidence does not inherently imply the typological validity of the rationales, it nevertheless demonstrates that Llama is moderately successful at generating consistent rationales \footnote{We highlight several failure cases in Appendix \ref{app:failure}.}. This aids in the explainability of predictions.

\section{Conclusion}

We investigated whether LLMs can perform typological feature prediction through reasoning over linguistic evidence. Our results suggest that while metadata-only prompting underperforms baselines, LLMs excel at integrating phylogenetic, geographic and typological evidence in ways that classical methods cannot. Simultaneously, providing more evidence helps LLMs perform more consistently across language resource levels and typological feature types. We further conducted a human annotation study, which supports our prompt design and finds that most LLM rationales are consistent with the evidence given, offering a step toward the explainable prediction systems called for by \citet{bjerva-2024-role}.

\section*{Limitations}

\paragraph{Task assumptions.} Our evaluation tests predictions only on entries that were already observed. However, prediction of the 87\% of entries that are truly missing from URIEL+ might represent a harder task, and may involve more atypical (language, feature) pairs. Moreover, while many typological features lie on a gradient rather than being categorical or even binary features \cite{levshina2023we}, our task inherits the binary assumption of knowledge bases such as URIEL and URIEL+. We frame this work as an exploratory study of LLMs as a prediction method, rather than a comprehensive evaluation of their typological knowledge.

Furthermore, our proposed method relies on evidence being available from neighbouring languages, and is thus tied to the data availability of the knowledge base used. We acknowledge that where no such evidence exists, our method is inapplicable; hence it should not be taken as a substitute for field-linguistic documentation.

\paragraph{Evaluation scale.} Due to the computational costs of language model inference, we perform prediction only on a limited subset of 1,200 examples. While this may not be wholly representative of performance on the whole dataset, and may be subject to noise, we mitigate this concern by ensuring equal representation of language resource levels and typological feature types in both test and validation splits. We however acknowledge that the aforementioned costs pose a substantial barrier to large-scale typological feature prediction or even database completion.

\paragraph{Explainability boundaries.} While our method generates rationales alongside predictions, we acknowledge that this may not be a truthful representation of the model's internal reasoning process \citep{10.5555/3666122.3669397}. Although the majority of rationales were found to be consistent, our failure analysis (Appendix \ref{app:failure}) confirms that correct predictions can occasionally accompany flawed rationales. Therefore, rationales are better understood as a plausible justification of the prediction only.

\paragraph{Model scope.} Our experiments cover two open-source model families alongside GPT-5.5, a proprietary frontier model. Hence, our evaluation remains limited to three model families. Nonetheless, the substantial improvements from linguistic evidence are observed across all three, suggesting that our findings are not necessarily tied to a single architecture or model family. Generally, our observed trends are stable across model families, suggesting that the central conclusions are not tied to a single architecture.

\paragraph{Data contamination.} URIEL+ does not contribute any novel typological data on its own, but rather aggregates multiple typological knowledge bases, which were released before the knowledge cut-off for the models used in our study. Thus, there is a minor risk that those models are relying only on pretrained knowledge. However, the poor performance under the base configuration across models, and the fact that URIEL+ additionally preprocesses the data in its constituent knowledge bases, suggests that data contamination poses only a minimal risk.

\section*{Ethical Considerations}

The task of typological feature prediction involves linguistic data from typological databases, which is free from any personally identifiable data. We source our data from URIEL+, which is publicly available.

We collect only numeric annotations of feature value predictions, confidences, and contradictions from human annotators in a structured file format. The data collected is thus free from personally identifiable information. All annotations are reported in aggregate, without any personal nor sensitive information.

\section*{Acknowledgments}
We are thankful to Richard Tzong-Han Tsai and his lab for helping run API-credentialed GPT-5.5 experiments. In addition, we are incredibly grateful to the following annotators for volunteering their time: Jiaqi Ji, Jinxi Li, Mason Shipton, and Qifan Yang. Finally, we thank Madelaine Jarcew for her feedback on the annotator study, and for providing us with linguistic resources to help annotators complete their task. 

\bibliography{main}
\bibliographystyle{acl_natbib}

\appendix

\section{ICL Prompt Template}
\label{app:prompt-template}

This appendix provides the prompt template used for in-context learning based on linguistic evidence.
\subsection{System Message}

\begin{quote}
You are a linguist working with data from a linguistic knowledge base.\\
Your task is to predict a target language's typological features given facts about the target language, other observed typological features, and evidence from phylogenetic and geographic neighbors.
\end{quote}

\subsection{User Prompt}

The user prompt is assembled from multiple evidence blocks, including metadata, anchor features, phylogenetic and geographic neighbours, vote summaries, contrastive nearest-neighbour evidence, global anchor-feature clues, and final output instructions.

\paragraph{Metadata.}
We include the language's name, Glottocode, ISO~639-3 code, family lineage (written as \texttt{family > parent} when both levels are distinct, as a single level otherwise, or as \texttt{Isolate}), macro-area(s) from Glottolog, and the name of the typological feature $f$ to predict.



\paragraph{Anchor features.}
To identify anchor features relevant to a target typological feature, we precompute a pairwise feature-correlation matrix from $\mathbf{M}_\text{train}$ using the Phi coefficient, following \citet{ng2025less}:
\[
\phi(f_i, f_j)=\frac{n_{11}n_{00}-n_{10}n_{01}}
{\sqrt{(n_{1\cdot}n_{0\cdot}n_{\cdot1}n_{\cdot0})}},
\]
where \(n_{ab}\) denotes the number of languages with feature values \(a\) for \(f_i\) and \(b\) for \(f_j\).

For a query language $\ell$, we include up to 5 observed anchor feature values. If fewer than 3 of the top-10 anchors are observed for $\ell$, the candidate pool is expanded to the top-15 correlated features before selecting up to 5 observed values.

We additionally compute, for each observed anchor feature $a$ with value $v_a$, the conditional support counts from $\mathbf{M}_\text{train}$, defined as the number of languages with $f\!=\!1$ and with $f\!=\!0$ given $a\!=\!v_a$ respectively. An aggregate tally then reports how many anchor features generally favour value 1, value 0, or are tied.

\paragraph{Phylogenetic neighbours.}
We retrieve 5 phylogenetic neighbours (i.e. languages with the exact same, or similar, family lineage) for $\ell$. We first build a candidate pool by breadth-first search over the Glottolog family tree starting from $\ell$, up to a limit of 400 languages. For language isolates, this pool is padded with arbitrary languages in a fixed order up to 400.

From this pool, neighbours are selected by a greedy procedure maximising coverage over the anchor and target features: at each step the candidate is chosen that covers the most anchor or target features not yet observed among selected neighbours, breaking ties by whether $f$ is itself observed, typological similarity to $\ell$, and proximity (i.e. shortest-path distance to $\ell$) rank. The nearest candidate in the full pool supporting \(f = 1\) is included by default beforehand, ensuring that contrastive evidence is always present.

For each selected neighbour, the prompt lists its observed value for $f$, followed by up to 3 observed anchor-feature values, together with its proximity rank in the full candidate pool. The block also includes:
\begin{itemize}
    \item A vote summary: counts of selected neighbours with $f\!=\!1$, $f\!=\!0$, and $f$ unobserved.
    \item The closest language in the full candidate pool (by genealogical proximity rank) supporting $f\!=\!1$, and the closest supporting $f\!=\!0$.
\end{itemize}

\paragraph{Geographic neighbours.}
Geographic neighbours are constructed identically, first constructing a candidate pool by ranking languages on Haversine distance from $\ell$'s Glottolog coordinates. Neighbour headers in the prompt report Haversine distance in km (rather than a proximity rank), as do the entries for the closest-supporting neighbours. The same greedy selection applies to geographic neighbours.

\paragraph{Task and output.}
The prompt asks the model to predict $f \in \{0,1\}$ and appends four reasoning guidance instructions: compare support for each value; weigh evidence holistically; note that a smaller number of closer neighbours may outweigh a larger but weaker group; and note that a minority value may be predicted if better supported. Output is constrained to one minified JSON object with keys \texttt{rationale} (at most 2 sentences), \texttt{value} (\texttt{"0"} or \texttt{"1"}), and \texttt{confidence} (\texttt{low}$|$\texttt{medium}$|$\texttt{high}). Three example outputs are appended in-prompt.

\subsection{Full Prompt Template}

\begin{tcolorbox}[
    enhanced,
    breakable,
    colback=gray!5,
    colframe=gray!50,
    arc=2mm,
    boxrule=0.6pt,
    left=2mm,
    right=3mm,
    top=2mm,
    bottom=2mm,
    before skip=4pt,
    after skip=4pt
]

\begin{Verbatim}[fontsize=\footnotesize]
Target language:
- Name: {lang_name}
- Glottocode: {glottocode}
- ISO639-3: {iso}
- Family lineage: {lineage}
- Macro-area: {macroarea}

Typological feature to predict:
{feature}

[Optional: anchor features]
The following are {lang_name}'s observed
typological features that correlate with 
the target feature:
- {anchor_feature_1}: {0|1}
- {anchor_feature_2}: {0|1}
- ...

[Optional: phylogenetic evidence]
The following languages are {lang_name}'s
phylogenetic neighbours (observations for 
anchor features and the target feature, 
when available, are listed):
1) {neighbor_name} (proximity rank={rank}):
   - {feature_or_anchor}: {0|1}
   - ...
2) ...

[Optional: geographic evidence]
The following languages are {lang_name}'s 
geographic neighbours (observations for 
anchor features and the targetfeature, 
when available, are listed):
1) {neighbor_name} (km={distance_km}):
   - {feature_or_anchor}: {0|1}
   - ...
2) ...

[Optional: vote summaries]
Phylogenetic neighbour votes:
- yes: {n_yes}
- no: {n_no}
- missing: {n_missing}

Geographic neighbour votes:
- yes: {n_yes}
- no: {n_no}
- missing: {n_missing}

[Optional: contrastive evidence]
The closest neighbours (by phylogeny and
geography) supporting each possible value:
- Closest phylogenetic neighbour supporting 
value 1:
  {label or none observed}
- Closest phylogenetic neighbour supporting 
value 0:
  {label or none observed}
- Closest geographic neighbour supporting 
value 1:
  {label or none observed}
- Closest geographic neighbour supporting 
value 0:
  {label or none observed}

[Optional: global anchor-feature clue 
summary]
Across the knowledge base, when languages 
exhibit the same anchor feature values:
1) {clue_feature}={0|1} ->
   {yes_count} languages support {feature}=1
   /{no_count} languages support {feature}=0
...

Overall:
{n_support_1} features support value 1 /
{n_support_0} features support value 0 /
{n_tied} features are tied

Task:
Predict the missing value for 
feature {feature} (allowed values: 0 or 1).

Reasoning guidance:
- Compare support for value 0 versus value 1.
- Weigh all evidence holistically.
- A smaller number of closer or more 
relevant neighbours may outweigh a larger 
but weaker group.
- A minority value may be predicted if 
better supported by the overall evidence.

Output:
Return exactly one valid minified JSON 
object on one line with keys:
- rationale
- value
- confidence

Constraints:
- rationale: at most 2 sentences
- value: "0" or "1"
- confidence: "low", "medium", or "high"

Example outputs:
{"rationale":"While the majority of 
neighbours support 0, the closest 
phylogenetic neighbours support 1.",
"value":"1","confidence":"medium"}

{"rationale":"Observed anchor features
and phylogenetic evidence strongly 
support value 1.","value":"1",
"confidence":"high"}

{"rationale":"Evidence is mixed, but anchor
statistics slightly favor value 0.",
"value":"0","confidence":"low"}
\end{Verbatim}

\end{tcolorbox}

\section{Test Set Composition}
\label{app:test-composition}

Table~\ref{tab:test-composition} provides additional statistics on the linguistic composition of the 1,200 test pairs. In addition to the balanced sampling across resource levels and feature types described in Section~\ref{sec:method}, the test set covers 171 distinct language families and a broad range of geographic macroareas.

\begin{table}[!htbp]
\centering
\small
\setlength{\tabcolsep}{5pt}
\begin{tabular}{p{0.29\columnwidth} p{0.63\columnwidth}}
\toprule
\textbf{Category} & \textbf{Coverage} \\
\midrule
Resource levels
& 400 pairs each for LRL / MRL / HRL \\

Feature types
& 300 pairs each for S / P / INV / M \\

Distinct families
& 171 \\

Largest families
& Austronesian 15.6\%, Atlantic-Congo 15.0\%, Indo-European 9.1\% \\

Macroareas
& Africa 26.1\%, Eurasia 23.0\%, Papunesia 22.4\%, North America 10.6\%, South America 10.2\%, Australia 4.0\% \\
\bottomrule
\end{tabular}
\caption{Composition of the 1,200-pair test set. Family and macroarea percentages denote the proportion of test pairs belonging to each group.}
\label{tab:test-composition}
\end{table}

\section{Evaluation Metrics for Imbalanced Data}
\label{app:imbalance}

Typological features can be highly imbalanced, with some features predominantly taking value 0 or 1. Our main experiments therefore report macro-F1 rather than accuracy and include a majority baseline. The majority baseline achieves a macro-F1 of 0.653, compared with 0.909 for the best LLM configuration, suggesting that the observed gains cannot be explained by majority-label prediction alone.

To further evaluate performance under label imbalance, we additionally report sensitivity, specificity, and balanced accuracy. Sensitivity measures recall on positive labels, specificity measures recall on negative labels, and balanced accuracy averages the two:
\[
\mathrm{BalAcc} = \frac{\mathrm{Sensitivity}+\mathrm{Specificity}}{2}.
\]

\newcommand{\bcell}[1]{%
  \cellcolor{blue!\fpeval{round(1 + (#1 - 0.311) / (0.986 - 0.311) * 50)}!white}#1%
}
\newcommand{\bcellb}[1]{%
  \cellcolor{blue!\fpeval{round(1 + (#1 - 0.311) / (0.986 - 0.311) * 50)}!white}\textbf{#1}%
}

\begin{table}[!htbp]
\centering
\small
\setlength{\tabcolsep}{3pt}
\begin{tabular}{
p{0.44\columnwidth}
p{0.2\columnwidth}
p{0.08\columnwidth}
p{0.08\columnwidth}
p{0.08\columnwidth}
}
\toprule
\textbf{Method} & \textbf{Inputs} & \textbf{Sens.} & \textbf{Spec.} & \textbf{Bal. Acc.} \\
\midrule

Random
& -- 
& \bcell{.600}
& \bcell{.785}
& \bcell{.692} \\

Majority
& --
& \bcell{.617}
& \bcell{.885}
& \bcell{.751} \\

kNN (cosine) ($k=3$)
& (A)
& \bcell{.786}
& \bcell{.898}
& \bcell{.842} \\

SoftImpute
& (A)
& \bcell{.823}
& \bcell{.924}
& \bcell{.873} \\

kNN (phylogenetic) ($k=6$)
& (A)+(B)
& \bcell{.775}
& \bcell{.882}
& \bcell{.828} \\

kNN (geographic) ($k=6$)
& (A)+(C)
& \bcell{.746}
& \bcell{.876}
& \bcell{.811} \\

Random Forest
& (A)+(B)+(C)
& \bcell{.792}
& \bcellb{.961}
& \bcell{.876} \\

\addlinespace
\multicolumn{5}{l}{\textit{Llama-3.1-70B in-context prompting}} \\

\quad Base, zero-shot \\
(metadata only)
& --
& \bcell{.792}
& \bcell{.322}
& \bcell{.557} \\

\quad + anchor features
& (A)
& \bcell{.907}
& \bcell{.311}
& \bcell{.609} \\

\quad + anchor + phylogenetic
& (A)+(B)
& \bcell{.955}
& \bcell{.795}
& \bcell{.875} \\

\quad + anchor + geographic
& (A)+(C)
& \bcell{.839}
& \bcell{.878}
& \bcell{.859} \\

\quad + all
& (A)+(B)+(C)
& \bcell{.961}
& \bcell{.936}
& \bcellb{.948} \\

\addlinespace
\multicolumn{5}{l}{\textit{Gemma-4-31B in-context prompting}} \\

\quad Base, zero-shot \\
(metadata only)
& --
& \bcell{.521}
& \bcell{.799}
& \bcell{.660} \\

\quad + anchor features
& (A)
& \bcell{.589}
& \bcell{.872}
& \bcell{.730} \\

\quad + anchor + phylogenetic
& (A)+(B)
& \bcell{.921}
& \bcell{.880}
& \bcell{.901} \\

\quad + anchor + geographic
& (A)+(C)
& \bcellb{.986}
& \bcell{.907}
& \bcell{.946} \\

\quad + all
& (A)+(B)+(C)
& \bcell{.980}
& \bcell{.879}
& \bcell{.930} \\

\bottomrule
\end{tabular}
\caption{Sensitivity, specificity, and balanced accuracy for the feature-prediction methods. Inputs follow Table~\ref{tab:main-results}: (A)=typological matrix, (B)=phylogenetic neighbours, and (C)=geographic neighbours. Cell shading is column-wise (darker = higher performance); per-column best is in \textbf{bold}.}
\label{tab:imbalance_metrics}
\end{table}

The imbalance-aware metrics preserve the main trend observed with macro-F1. Metadata-only and anchor-only prompting show substantial asymmetry between positive- and negative-label performance, particularly for Llama-3.1-70B. Performance becomes considerably more balanced once neighbour evidence is introduced.

Among the non-LLM baselines, Random Forest and SoftImpute achieve balanced accuracies of 0.876 and 0.873, respectively. In comparison, Llama-3.1-70B with all evidence achieves the highest balanced accuracy of 0.948, with sensitivity 0.961 and specificity 0.936. Gemma-4-31B with anchor and geographic evidence performs similarly, reaching balanced accuracy 0.946 with sensitivity 0.986 and specificity 0.907. These results indicate that the strongest evidence-grounded LLM configurations perform well on both labels rather than obtaining their gains primarily from dominant feature values.

\section{Comparison Against a Plurality Voting Heuristic}
\label{app:plurality}
A reasonable concern about the performance of LLMs is whether they are applying any reasoning beyond the heuristic of simply following the plurality vote from neighbours or anchor features. The random forest baseline partly addresses this, as it is trained over the same evidence the LLM sees, and would capture such a rule if one sufficed for the task; yet the random forest model reaches only 0.731 macro F1, falling far short of Llama-3.1-70B's 0.909.

To more rigorously test how far the support counts alone determine the gold value, we evaluate plurality voting over each evidence block on its own, predicting the value supported by the majority of observed phylogenetic neighbours, geographic neighbours, or anchor features respectively. Where there is a tie, a value of 1 is predicted. We additionally test these methods on a non-trivial subset of test examples, comprising 10.2\% of the test set, where phylogenetic and geographic plurality votes disagree.

\begin{table}[t]
\centering
\small
\begin{tabular}{lcc}
\toprule
Method & Full & Non-trivial \\
\midrule
Llama-3.1-70B & \textbf{.909} & \textbf{.893} \\
Phylogenetic neighbours, plurality & .770 & .538 \\
Geographic neighbours, plurality & .759 & .459 \\
Anchor features, plurality & .735 & .518 \\
\bottomrule
\end{tabular}
\caption{Macro F1 of Llama-3.1-70B under the all-inputs configuration against plurality voting over individual evidence blocks, both on the full test set and on the non-trivial subset.}
\label{tab:plurality}
\end{table}

Table~\ref{tab:plurality} shows the performance of plurality voting compared to Llama-3.1-70B on both the full test set and the non-trivial subset. We observe that under all evidence blocks, simply predicting based on the plurality vote yields a performance substantially lower than the LLM. This gap is particularly stark on the non-trivial subset, where plurality voting underperforms even further, whereas LLM prediction retains a respectable macro F1 of 0.893. This therefore shows that the LLMs' aggregation of evidence is not reducible to simply counting votes, and in fact LLMs hold an advantage where the evidence conflicts.

\section{Within-Family Model Size Comparison}
\label{app:model-size}

\begin{table}[h]
\centering
\small
\setlength{\tabcolsep}{8pt}
\begin{tabular}{l c c}
\toprule
\textbf{Config} & \textbf{Llama 8B} & \textbf{Llama 70B} \\
\midrule
Base (metadata only)
& \cellcolor{blue!0}.147
& \cellcolor{blue!0}.465 \\

\,+\,anchor features
& \cellcolor{blue!18}.425
& \cellcolor{blue!2}.512 \\

\,+\,anchor\,+\,phylogenetic
& \cellcolor{blue!23}.480
& \cellcolor{blue!27}.782 \\

\,+\,anchor\,+\,geographic
& \cellcolor{blue!45}\textbf{.661}
& \cellcolor{blue!28}.788 \\

\,+\,all
& \cellcolor{blue!30}.535
& \cellcolor{blue!45}\textbf{.909} \\

\bottomrule
\end{tabular}
\caption{Overall macro F1 for Llama-3.1 at two model sizes. Cell shading is column-wise (darker = higher F1).}
\label{tab:llama-scale}
\end{table}

To study the effect of model size on imputation performance, Table \ref{tab:llama-scale} shows the macro F1 scores for Llama-3.1 at 8B and 70B sizes. Across configurations, we find that performance increases as model size increases, with a particularly substantial increase at the base configuration (F1 0.147 vs 0.465). This is consistent with our expectation that larger models have more intrinsic knowledge (limited as it is) and better reasoning abilities. 
Interestingly, the 8B model exhibits a similar trend to Gemma-4-31B, with performance greater under geographic evidence than under the configuration containing all inputs. GPT-5.5 exhibits the same pattern in Table~\ref{tab:main-results}. This suggests that the effect is not explained by simply the model size. It is the case that heterogeneous evidence can sometimes reduce performance.

\section{Annotation Details and Protocol}\label{app:annotation-protocol}
Five annotators were recruited for the annotation task described in Section \ref{sec:human}. Annotators were undergraduate students with backgrounds in computer science or linguistics, from the authors' institution. They were informed that their annotation would be used for a research context (along with contextual information about what the task would entail), and that they would participate voluntarily without monetary compensation, to which they provided their explicit consent.

For each of the 180 $(\ell, f)$ cases collected according to the rule stated in Section \ref{sec:human}, the annotation task proceeded in two stages.
\begin{enumerate}
    \item \textbf{Stage 1:} Given the same typological facts given to the LLM (namely, all four context blocks described in Section \ref{sec:prompt}), the annotator was asked to decide whether the target language has the typological feature stated or not (1/0). In addition, they were asked to state their confidence level on the same scale as the LLM (High/Medium/Low). They were also told to mark their \emph{absolute confidence} in their prediction, rather than their \emph{relative confidence} between cases.
    \item \textbf{Stage 2:} Given the exact same set of cases as in Stage 1, the annotator is now presented with what Llama-3.1-70B predicted, as well as their rationale. Their job was to determine whether the rationale of given by the LLM is consistent with the information provided to them, and the prediction they made (Yes/No). Note that Llama-3.1-70b was chosen, due to its superior performance in Table \ref{tab:main-results}.
\end{enumerate}
Each annotator completed the task independently of each other. These cases were presented to annotators in a well-designed web app for ease of annotation. Within a particular stage, annotators completed each case sequentially, with the option to go back to a previous case if desired. However, annotators were instructed to only start Stage 2 once they completed Stage 1. This was to avoid their decision making in Stage 1 from being influenced by the LLM's reasoning process (which Stage 2 exposes). 

To aid in their decision-making, we provided two aid sheets: 1) an IPA chart\footnote{IPA Kiel, revised to 2015}, and 2) a two-page reference sheet on linguistics concepts containing information on phonetics, phonology, morphology, and syntax, and how to read the IPA chart. Annotators were strictly forbidden from using LLMs or AI tools for this task, or on any material related to this task. 

Annotators were allowed as much time as desired to complete this task, but we informed them that we expect each case per stage, to take roughly three minutes. Annotators shared their results with the investigators independently of each other.

\section{Qualitative Analysis of LLM Rationales}\label{app:failure}

In this section, we provide a qualitative analysis of Llama-3.1-70B rationales to understand what evidence the model appears to use, when this evidence use leads to correct predictions, and when it produces incomplete or misleading explanations. In particular, from the 180 cases used in the annotation study, we construct two subsets of rationales to analyse. 

The first subset consists of all cases where a majority of annotators (i.e. three) agreed that the LLM provided a flawed rationale. There were 30 such majority-contradiction cases. On this restricted subset, the LLM still achieved accuracy 0.867 (F1 0.464), while the human plurality achieved accuracy 0.967 (F1 0.492). The low F1 values mainly reflect the strong class imbalance in this subset, where almost all gold labels are 0. Only four cases combined a majority-marked inconsistent rationale with an incorrect LLM prediction. We call this the \emph{contradiction subset}. 

To match this subset, we also randomly sampled another 30 cases where no annotator marked it as having a reasoning error. On this subset, the LLM predicted 29 of 30 cases correctly. We call this the \emph{non-contradiction subset}. 

Of note, in \emph{none} of the cases studied, do the rationales mention the target language's metadata, or any external fact about the language not provided by the prompt. Instead, the rationales overwhelmingly cite evidence explicitly present in the prompt (or no evidence at all). Thus, the model's successful explanations are best understood as prompt-grounded evidence summaries, not as demonstrations of innate typological knowledge.

\paragraph{Pattern 1: Cross-source agreement produces the clearest successes.}
The clearest successful rationales occur when phylogenetic neighbours, geographic neighbours, and anchor-feature statistics all suggest the same label. For example, \texttt{dalo1238} (Daloa B\'et\'e), a medium-resource language with target feature \texttt{P\_BILABIALS}, had unanimous neighbour support for value 1: 5/5 phylogenetic neighbours and 5/5 geographic neighbours supported 1, and both available anchor clues also supported 1. The LLM predicted the gold value 1 with high confidence and gave the rationale: \emph{``The observed anchor features are known to support 1 when they are 0. Furthermore, both phylogenetic and geographic neighbours align strongly with value 1, with all observed neighbours supporting 1.''} This is a faithful summary of the prompt evidence.

With that said, the contradiction subset shows that even when all the evidence in the prompt is aligned, the model can still fail to report it. For \texttt{shan1277} (Shan), a low-resource language with target feature \texttt{P\_LATERAL\_OBSTRUENTS}, both phylogenetic and geographic neighbours gave majority support for value 0, and the anchor clue also supported value 0. The model nevertheless predicted 1 and gave only the rationale \emph{``Insufficient direct evidence''}. Even if the evidence was given to the model, it failed to interpret it, and provided an incorrect prediction. 

\paragraph{Pattern 2: Neighbour evidence is usually primary, while anchor evidence is usually secondary.}
Another observation is that the LLM usually follows neighbour evidence, especially when phylogenetic and geographic neighbours agree. A successful example from the non-contradiction set is \texttt{awad1243} (Awadhi), a high-resource language with target feature \texttt{M\_PRESENT\_MARK}. The single anchor clue supported value 0, but neighbour evidence strongly supported value 1: four of five phylogenetic neighbours and all five geographic neighbours supported 1. The LLM predicted the gold value 1 with high confidence, stating that \emph{``the majority of phylogenetic and geographic neighbours support value 1''} and that the closest neighbours also supported 1, \emph{``outweighing the general trend''} from languages with the same anchor value. Hence it prioritised local neighbour evidence over a global anchor trend, rather than a prediction based on independent knowledge about Awadhi's family or typological profile.

Even if the model appears to favour neighbour evidence, it does not always make it clear how it weighs differing neighbour information. Indeed, from the contradiction subset, we may consider the case of \texttt{dura1244} (Dura). This is a medium resource language with target feature \texttt{S\_POLARQ\_MARK\_SECOND}. The majority of neighbours supported 0, but the closest phylogenetic and geographic neighbours both supported 1. The model predicted correctly, but its rationale only began to describe the conflict: \emph{``Although the majority of both phylogenetic and geographic neighbours support 0, the closest neighbour in both categories, Bujhyal, supports 1''}. It did not give an explanation of how it weighed the neighbour evidence given. 

A similar, but distinct observation is that for low-resource cases, the LLM had a reliance on using neighbour evidence, according to its rationale. This is generally because these cases often provided fewer target-language anchor features, which when combined with the fact that the LLM naturally seems to disfavour anchor information, resulted in this reliance.  

\paragraph{Pattern 3: Proximity and contrastive evidence are useful, but can be incorrectly interpreted.}
Recall that the prompt asks the model to consider whether a smaller number of closer neighbours may outweigh a larger but weaker group. From the non-contradiction subset, we see that the LLM can use this instruction sensibly. For example, \texttt{chak1270} (Chak), a medium-resource language with target feature \texttt{M\_COME\_VERB\_SUPPLETION}, had four of five phylogenetic neighbours and four of five geographic neighbours supporting 0. The closest geographic neighbour supporting 0 was Anu-Hkongso at 48.0 km, while the closest geographic neighbour supporting 1 was Zeme Naga at 450.7 km. The LLM predicted the gold value 0 with medium confidence and explained: \emph{``The majority of both phylogenetic and geographic neighbours support value 0, and the closest geographic neighbour supporting value 0 is significantly closer than the closest geographic neighbour supporting value 1.''} 

The contradiction cases show that proximity evidence can be misinterpreted or even omitted from reasoning though. In \texttt{anut1237} (Anuta), a low-resource language with target feature \texttt{M\_UNPREDICTABLE\_GENDER\_CLASS}, the prompt gave no target-language anchor features, so the model had to rely on neighbour evidence. The rationale stated that \emph{``the majority of phylogenetic and geographic neighbours support value 0''} and noted that the closest phylogenetic neighbour supporting 0 was more proximal than the closest phylogenetic neighbour supporting 1. This correctly described part of the evidence, and the model predicted the gold value. However, the rationale omitted the countervailing geographic information: the closest geographic neighbour supporting 1 was closer than the closest geographic neighbour supporting 0. In \texttt{nata1254} (Northern Amis), a medium-resource language with target feature \texttt{M\_PROD\_PLURAL\_MARK}, the model similarly cited closest-neighbour support for value 0, saying that \emph{``the closest phylogenetic neighbour Sirayaic and geographic neighbour Atayal support value 0''}, but then added that \emph{``the observed anchor features do not provide strong evidence for value''}, even though the prompt did not provide target-language anchor features. These are not necessarily prediction failures, but they are rationale-faithfulness failures: the model gives a simplified proximity account that omits or misstates part of the prompt evidence.

Proximity can also be overweighted. The one incorrect prediction in the no-contradiction subset was \texttt{garo1247} (Garo), a high-resource language with target feature \texttt{P\_IMPLOSIVES}. The gold label and human majority were 0. Both neighbour blocks supported 0 by a 4-to-1 majority, and both anchor clues supported 0. The LLM nevertheless predicted 1 with medium confidence. Its rationale began: \emph{``Although the majority of phylogenetic and geographic neighbours support 0, the closest phylogenetic and geographic neighbours supporting value 1 are relatively closer than those supporting value 0.''} This rationale overstates the importance of proximity. Indeed, the positive phylogenetic supporter was only slightly closer than the nearest negative supporter, while the nearest geographic supporter of 0, Assamese at 121.2 km, was much closer than the nearest geographic supporter of 1, Nepali Kurux at 473.0 km. Thus, proximity is useful when it reinforces broader evidence, but it can become misleading if the evidence is not given appropriate weight.

\paragraph{Pattern 4: Generic rationales often mark uncertainty.}
One fallback phrase that the LLM provided in both non-contradiction and contradiction subsets was: \emph{``Insufficient direct evidence''}. In the no-contradiction subset, four of the 30 rationales used exactly this phrase; all four predictions were correct and low-confidence. Hence, it seems to be an indicator for a lack of confidence on the model's part. 

In the contradiction subset, the phrase occurred in 8 of the 30 majority-contradiction cases, and all four cases that combined a majority-marked inconsistent rationale with an incorrect LLM prediction used this fallback rationale. One can recall a case with Shan, where it ignored aligned support throughout the prompt for 0, accompanied by rationale. Another case is \texttt{ewee1241} (Ewe), a high-resource language with target feature \texttt{M\_PROD\_AUGMENTATIVE\_NOUN}, the same rationale was given in a conflicted case: phylogenetic neighbours and anchor clues supported 0, while geographic neighbours strongly supported 1. The gold label was 1, but both the LLM and the human majority predicted 0. Here the problem was that the rationale failed to describe the conflict that made the case difficult.

\paragraph{Pattern 5: Phonological and inventory cases are hard with conflicting evidence.}
The model can handle phonological and inventory features when evidence is clear. An example is \texttt{dalo1238} (Daloa Bété) for \texttt{P\_BILABIALS}. However, the majority-contradiction subset suggests that fine-grained segmental features are a common source of actual errors when evidence conflicts. Of the four cases where a majority-marked inconsistent rationale coincided with an incorrect prediction, three involved phonological or inventory features. In \texttt{nort2921} (North Wahgi), a low-resource language with target feature \texttt{P\_FRICATIVES}, the neighbour majority supported value 0, but the anchor clue supported 1; the model gave the generic rationale \emph{``Insufficient direct evidence''} and predicted 1, while both the gold label and human majority were 0. The aforementioned Shan and Garo cases are similar. These cases suggest that the model's evidence weighting for these kinds of cases is brittle, when the evidence given can be difficult to interpret.

\paragraph{The takeaway.}
While the rationales given by the LLM is a step towards explainable typological prediction, they are best understood as a method to interpret evidence given to it. From the rationales analysed, we do not have any reason to say that the LLM uses any innate typological knowledge, or knowledge about the language outside of the prompt (despite given metadata) to reason in this task. Furthermore, the LLM appears to have innate biases in how it prioritises certain information over others. Nonetheless, even if the rationale the LLM provides sometimes does not provide a full account of its behaviour, it does offer (often strong) clues towards its prediction.

\section{Baseline Architectures}\label{app:hparams}

\subsection{k-Nearest Neighbours}

We chose $k=3$ for \texttt{kNN-cosine} and $k=6$ for \texttt{kNN-phylogenetic} and \texttt{geographic}. This was determined following a search over $k$, evaluating each model on macro F1 on the validation split. We then choose the value of $k$ where performance begins to plateau or decrease. Where a tie occurs, a value of $1$ is predicted.

\subsection{Random Forest Classifier}

Following \citet{amirzadeh2025data2lang2vec}, we train a random forest classifier for each typological feature $f$. To best match the inputs visible to the LLM, we train the classifier on the following features given some language $\ell$ (totalling 42 features):

\paragraph{Metadata}
\begin{itemize}
\item $P(f=1|\,\text{top-level family})$
\item $P(f=1|\,\text{macroarea})$
\end{itemize}

\paragraph{Anchor features}
For each of the top 5 correlated anchor features $a$ which are observed for $\ell$:
\begin{itemize}
    \item $\ell$'s value of feature $a$, $v_a=\textbf{M}_\text{train}[\ell,a]$
    \item Global anchor feature clue, $P(f=1|\,a=v_a)$
\end{itemize}

\paragraph{Phylogenetic neighbours}
For each of the 5 nearest phylogenetic neighbours $\ell_p$:
\begin{itemize}
    \item $\ell_p$'s value of $f$, $\textbf{M}_\text{train}[\ell_p,f]$
    \item Whether $f$ is missing in $\ell_p$
\end{itemize}
Supplemented by the following aggregate features:
\begin{itemize}
    \item Number of phylogenetic neighbours voting yes (i.e. where $\textbf{M}_\text{train}[\ell_p, f]=1 \land \textbf{M}_\text{train}[\ell_p, f]\neq \bot$)
    \item Number of phylogenetic neighbours voting no (i.e. where $\textbf{M}_\text{train}[\ell_p, f]=0 \land \textbf{M}_\text{train}[\ell_p, f]\neq \bot$)
    \item Number of phylogenetic neighbours with missing $f$
    \item The rank of the closest phylogenetic neighbour voting yes
    \item The rank of the closest phylogenetic neighbour voting no
\end{itemize}

\paragraph{Geographic neighbours}
For each of the 5 nearest geographic neighbours $\ell_g$:
\begin{itemize}
    \item $\ell_g$'s value of $f$, $\textbf{M}_\text{train}[\ell_g,f]$
    \item Whether $f$ is missing in $\ell_g$
\end{itemize}
Supplemented by the following aggregate features:
\begin{itemize}
    \item Number of geographic neighbours voting yes (i.e. where $\textbf{M}_\text{train}[\ell_g, f]=1 \land \textbf{M}_\text{train}[\ell_g, f]\neq \bot$)
    \item Number of geographic neighbours voting no (i.e. where $\textbf{M}_\text{train}[\ell_g, f]=0 \land \textbf{M}_\text{train}[\ell_g, f]\neq \bot$)
    \item Number of geographic neighbours with missing $f$
    \item The rank of the closest geographic neighbour voting yes
    \item The rank of the closest geographic neighbour voting no
\end{itemize}

Following a grid search over the number of estimators \texttt{n\_estimators} and maximum tree depth \texttt{max\_depth}, evaluating macro F1 performance on the validation split, we choose \texttt{n\_estimators}$=500$ and  \texttt{max\_depth}$=8$, with the remaining hyperparameters set to scikit-learn \citep{scikit-learn} defaults.

\section{Computing Infrastructure}

Computational experiments on all three open-source models (Llama-3.1-8B, Llama-3.1-70B, Gemma-4-31B) across all five configurations were run on a single Nvidia H100 GPU, totalling 25 compute hours. GPT-5.5 experiments were conducted through inference on Azure OpenAI. All other experiments were run on CPU.

\begin{table*}[h]
\centering
\begin{tabular}{ll}
\toprule
\textbf{Artifact} & \textbf{License} \\
\midrule
\multicolumn{2}{l}{\textit{Packages}} \\
URIEL+ \citep{khan2025uriel+} & CC BY-SA 4.0 \\
scikit-learn (v1.8) \citep{scikit-learn} & BSD-3-Clause \\
Transformers (v5.1) \citep{wolf-etal-2020-transformers} & Apache 2.0 \\
\midrule
\multicolumn{2}{l}{\textit{Datasets}} \\
Glottolog (v5.2) \citep{Glottolog} & CC BY 4.0 \\
\midrule
\multicolumn{2}{l}{\textit{Models}} \\
Llama-3.1-8B-Instruct \citep{grattafiori2024llama} & Llama 3.1 Community License Agreement \\
Llama-3.1-70B-Instruct \citep{grattafiori2024llama} & Llama 3.1 Community License Agreement \\
Gemma-4-31B-IT \citep{gemma4} & Apache 2.0 \\
GPT-5.5 \citep{openai2026gpt55} & Azure OpenAI (Microsoft Product Terms) \\
\bottomrule
\end{tabular}
\caption{Artifacts used in this study, and their licenses.}
\label{tab:artifacts}
\end{table*}


\section{Use of Generative AI}
We used generative AI only in a limited capacity. Namely, generating auto-code completions (which were verified) and checking the grammar and structure of our text. 

\section{Licenses for Artifacts}
We present the licenses for this study in Table \ref{tab:artifacts}. Our usage of these artifacts is in line with the licenses these artifacts were released under.

\end{document}